%% file: preprint_arXiv_2026.tex
\documentclass{article}

\PassOptionsToPackage{numbers, compress}{natbib}
\usepackage[preprint]{neurips_2026}

\usepackage[utf8]{inputenc} % allow utf-8 input
\usepackage[T1]{fontenc}    % use 8-bit T1 fonts
\usepackage{hyperref}       % hyperlinks
\usepackage{url}            % simple URL typesetting
\usepackage{booktabs}       % professional-quality tables
\usepackage{amsfonts}       % blackboard math symbols
\usepackage{nicefrac}       % compact symbols for 1/2, etc.
\usepackage{microtype}      % microtypography
\usepackage{xcolor}         % colors
\usepackage{times}
\usepackage{graphicx}
\usepackage{booktabs}
\usepackage{subcaption}
\usepackage{hyperref}
\usepackage{float}
\usepackage{enumitem}

\usepackage{amsmath, amssymb, mathtools}
\usepackage{natbib}
\usepackage{booktabs}   % For professional lines
\usepackage{xcolor}     % For color
\usepackage{colortbl}   % For colored rows
\usepackage[utf8]{inputenc}
\usepackage{booktabs} % For professional table lines
\usepackage{tabularx} % For adjustable width tables

\usepackage[ruled,vlined,linesnumbered]{algorithm2e}
\usepackage{xcolor}
\definecolor{commentgray}{rgb}{0.5,0.5,0.5}

\SetCommentSty{mycommfont}

\title{Multi-Objective Constraint Inference using Inverse reinforcement learning}

\author{%
  Syed Ihtesham Hussain Shah \\
  Faculty of Sciences \\
  Vrije Universiteit Amsterdam\\
  \texttt{s.i.h.shah@vu.nl} \\
   \And
   Floris den Hengst \\
   Faculty of Sciences \\
  Vrije Universiteit Amsterdam\\
  \texttt{f.den.hengst@vu.nl} \\
   \AND
   Aneta Lisowska \\
    Faculty of Sciences \\
  Vrije Universiteit Amsterdam\\
  \texttt{a.j.lisowska@vu.nl} \\
  \And
  Annette ten Teije \\
    Faculty of Sciences \\
  Vrije Universiteit Amsterdam\\
  \texttt{annette.ten.teije@vu.nl} \\
}

\begin{document}

\maketitle

\begin{abstract}
Constraint inference is widely considered essential to align reinforcement learning agents with safety boundaries and operational guidelines by observing expert demonstrations. However, existing approaches typically assume homogeneous demonstrations (i.e., generated by a single expert or multiple experts with identical objectives). They also have limited ability to capture individual preferences and often suffer from computational inefficiencies.
In this paper, we introduce Multi-Objective Constraint Inference (MOCI), a novel framework designed to jointly extract shared constraints and individual preferences from heterogeneous expert trajectories, where multiple experts pursue different objectives. MOCI effectively models and learns from diverse, and potentially conflicting, behaviors.
Empirical evaluations demonstrate that MOCI significantly outperforms existing baselines, achieving improved predictive performance, and maintaining competitive computational efficiency on a standard grid-world benchmark. These results establish MOCI as an accurate, flexible, and computationally practical approach for real-world constraint inference and preference learning tasks.
\end{abstract}
%%%%%%%%%%%%%%%%%%%%%%%%%%%%%%%%
\input{Sections/Introduction}
\input{Sections/Prelaminaries}
\input{Sections/Approach}
\input{Sections/ComputationCost}
\input{Sections/ExperimentAndResults}
\input{Sections/Comparison_with_Baseline}
\input{Sections/Conclusion}
%%%%%%%%%%%%%%%%%%%%%%%%%%%%%%

\noindent \textbf{Acknowledgments} This work has been partly supported by the PersOn project (P21-03) and the Hybrid Intelligence center (grant no. 024.004.022), both funded by Nederlandse Organisatie voor Wetenschappelijk Onderzoek (NWO).

\bibliographystyle{unsrtnat}
\bibliography{references}
\newpage
\appendix
\input{Sections/Apendics}
%
%%%%%%%%%%%%%%%%%%%%%%%%%%%%%%%%%%%%%%%%%%%%%%%%%%%%%%%%%%%%
% %
% %
% \section{Technical appendices and supplementary material}
% Technical appendices with additional results, figures, graphs, and proofs may be submitted with the paper submission before the full submission deadline (see above). You can upload a ZIP file for videos or code, but do not upload a separate PDF file for the appendix. There is no page limit for the technical appendices. 

% Note: Think of the appendix as ``optional reading'' for reviewers. The paper must be able to stand alone without the appendix; for example, adding critical experiments that support the main claims to an appendix is inappropriate. 
\newpage
% %%%%%%%%%%%%%%%%%%%%%%%%%%%%%%%%%%%%%%%%%%%%%%%%%%%%%%%%%%%%

%\input{checklist.tex}

\end{document}

%% file: Sections/Introduction.tex
\section{Introduction}\label{sec:intro}
Inverse Reinforcement Learning (IRL) is a framework for learning the underlying objectives of an expert by observing their behavior \cite{shah2022learning}. Instead of manually specifying a reward function, IRL infers the reward that best explains the expert's demonstrated actions.
Traditionally, the intent of an agent is modeled by a learned reward function, however, accurately matching the expert's behavior can make the reward overly complex and brittle to small changes. Instead of capturing complex and diverse motivations of the expert in a single reward function, expert behavior can be more easily explained by jointly learning a simple reward function and inferring a set of clear hard constraints \cite{scobee2019maximum}.
\\
Although jointly learning constraints and reward functions yield simpler reward models that better explain expert trajectories, existing work \cite{mcpherson2021maximum, malik2021inverse} suffers from two restrictive assumptions: (1) the dataset is homogeneous, i.e. generated by a single type of expert and (2) the expert's reward function is completely known \textit{a priori}. 
These assumptions are problematic in real-world settings where observed data typically emerge from a heterogeneous population of experts, and the reward function is not known. For instance, in urban driving, different drivers exhibit varying driving styles, ranging from aggressive to cautious, yet all are subject to the same shared physical and legal constraints, such as speed limits and lane boundaries. At the same time, drivers have different driving and navigation preferences such as reaching the destination with minimum distance, or taking a longer route with more pleasant sceneries. 
\\
In this paper, we address the challenge of inferring shared constraints and learning individual preferences from heterogeneous expert demonstrations. We propose Multi-Objective Constraint Inference (MOCI), a novel framework to jointly recover shared constraints and preferences from unlabeled demonstrations generated by multiple experts with varying preferences and in an environment with hard constraints shared across all experts.
\\
MOCI iteratively clusters expert trajectories while learning personalized reward weights. Based on these clusters (groups), MOCI then identifies shared constraints $C$ by evaluating each state using the joint log-likelihood method \cite{zeng2022maximum}. A hard constraint is identified for a state  only if its removal from the feasible environment increases the likelihood of the observed demonstrations across all clusters simultaneously. Hence, MOCI jointly learns shared constraints and personalized reward weights by applying maximum entropy inverse reinforcement learning to each group, allowing the separation of individual expert preferences from shared environmental constraints.
\\
We evaluate our approach on a multi-objective GridWorld environment, a well-established benchmark for sequential decision-making problems. This controlled setting allows us to systematically validate the proposed method.
\\
The remainder of this paper is organized as follows: Section \ref{sec:preliminaries} formalizes the preliminaries and foundational concepts upon which the framework is built, including Constrained Markov Decision Processes (CMDPs) and Maximum Entropy Inverse Reinforcement Learning (IRL). Section \ref{sec:methods} introduces the proposed Multi-Objective Constraint Inference (MOCI) algorithm and provides a detailed analysis of its computational complexity. Section \ref{sec:experiments} details the experimental setup using a simulated heterogeneous Gridworld environment and discusses the empirical results of the framework. Section \ref{Sec:comparison} provides a performance comparison between MOCI and existing baseline techniques. Finally, Section \ref{Sec:conclusion} summarizes the study's conclusions and discusses current limitations.

%% file: Sections/Prelaminaries.tex
%%%%%%%%%%%%%%%%%%%%%%%%
\section{Preliminaries}
\label{sec:preliminaries}
%%%%%%%%%%%
In this section, we formalize the foundational concepts upon which the Multi-Objective Constraint Inference (MOCI) framework is built. We define the environments, the structure of the agent's objectives, and the methodologies used to infer those objectives from the data.
%%%%%%%
\subsection{Constrained Markov Decision Processes}
%%%%%%
A standard Markov Decision Process (MDP) is defined as a tuple $\mathcal{M} = \langle \mathcal{S}, \mathcal{A}, \mathcal{T}, \gamma, R \rangle$, where $\mathcal{S}$ is the state space, $\mathcal{A}$ is the action space, $\mathcal{T}(s' \mid s, a) \in \mathcal{S} \times \mathcal{A} \to \Delta(\mathcal{S})$ is the transition probability distribution, $\gamma \in [0, 1)$ is the discount factor, $R(s, a) \in \mathcal{S} \times \mathcal{A} \to \mathbb{R}$ is the reward function \cite{singh2022learning}. The behavior of an agent is defined by a policy $\pi(a \mid s) \in \mathcal{S} \to \Delta(\mathcal{A})$, which assigns states to a probability distribution over actions.
\\
A \textit{Constrained Markov Decision Process} (CMDP) \cite{wachi2020safe} extends this framework by restricting the set of allowable policies. Although CMDP can be formulated using cost functions and budget thresholds, we focus on the context of hard environmental constraints (such as walls or strict physical limitations), defined as a set of forbidden or unsafe states denoted $C \subset \mathcal{S}$ following recent related work by~\citet{kim2023learning,qadri2025your}. 
\\
The CMDP is thus augmented to $\mathcal{M}_C = \langle \mathcal{S}, \mathcal{A}, \mathcal{T}, \gamma, R, C \rangle$. 
\\
A trajectory $\xi = \{(s_0, a_0), (s_1, a_1), \dots, (s_T, a_T)\}$ of length $H\in \mathbb{N}$ is considered valid if and only if it does not violate any constraints in $C$ \cite{schlaginhaufen2023identifiability}. This is formalized using a binary indicator function:
\begin{equation}
    \mathbb{I}^C(\xi) = 
    \begin{cases} 
      1 & \text{if } s_t \notin C \text{ for all } t \in \{0, \dots, H\} \\
      0 & \text{otherwise}
    \end{cases}
    \label{equ1}
\end{equation}

\subsection{Multi-Objective MDP}

A \textit{Multi-Objective Markov Decision Process} (MOMDP) extends traditional MDPs with a vectorial reward function $\boldsymbol{R} \in \mathcal{S} \times \mathcal{A} \to \mathbb{R}^d$ to model environments with $d$ potentially competing objectives (e.g., minimizing travel time while maximizing safety) \cite{barrett2008learning,hayes2022practical}. This vectorial reward can be collapsed into a scalar reward if the particular way in which these multiple objectives are to be combined for some particular individual or use case, are known. In case this so-called \emph{scalarization} function is linear in the objectives, we refer to its weights $w:=\left[w_1, \dots, w_p\right]$ as the preferences of that individual such that their reward function $R = w^\top \boldsymbol{R}$.
%\cite{delgrange2020simple}.
% In an MOMDP, the scalar reward function $R(s, a)$ is replaced by a vector-valued reward function $\mathbf{R}(s, a) \in \mathbb{R}^d$:
% \begin{equation}
%     \mathbf{R}(s, a) = [R_1(s, a), R_2(s, a), \dots, R_d(s, a)]^\top
% \end{equation}

In practice, this is often expressed via a feature formulation where $\phi(s, a) \in \mathbb{R}^d$ represents a vector of state-action features, and an agent's specific preference is defined by a weight vector $w$. The scalarized reward \cite{shah2023projection} for an agent with preference $w$ can be defined as:
\begin{equation}
    R_w(s, a) = w^\top \phi(s, a)
\end{equation}
In a heterogeneous multi-agent setting \cite{zhong2024heterogeneous}, different experts $k \in \{1, \dots, K\}$ share the same state-action features $\phi$ but possess distinct, private preference weights $w_k$, leading to diverse optimal policies within the same underlying environment.
%%%%%%%%%%
\subsection{Inverse Reinforcement Learning}
%%%%%%%%%%
\textit{Inverse Reinforcement Learning} (IRL) addresses the problem of extracting an agent's underlying reward function given its demonstrated behavior \cite{arora2021survey}. Formally, we assume access to the MDPs transition function but not its reward function, denoted as $\mathcal{M} \setminus R$, and a dataset of expert trajectories $\mathcal{D} = \{\xi_1, \dots, \xi_N\}$ generated by an expert policy $\pi_E$.

The goal of IRL is to find a reward function $R^*$ such that the expert's policy $\pi_E$ is optimal\cite{adams2022survey}. If the reward is parameterized linearly as $R(s, a) = w^\top \phi(s, a)$, the IRL problem can be framed as finding a weight vector $w^*$ such that the expected feature counts of the expert match the expected feature counts of a policy optimizing $w^*$:
\begin{equation}
    \mathbb{E}_{\pi_E}\left[\sum_{t=0}^T \gamma^t \phi(s_t, a_t)\right] = \mathbb{E}_{\pi_{w^*}}\left[\sum_{t=0}^T \gamma^t \phi(s_t, a_t)\right]
\end{equation}
However, this problem is inherently ill-posed, as multiple reward functions (including a trivial reward of all zeros) can explain the same behavior. 

\subsection{Maximum Entropy IRL}

To resolve the ambiguity of the IRL problem, \cite{ziebart2008maximum} introduced \textit{Maximum Entropy Inverse Reinforcement Learning} (MaxEnt IRL). MaxEnt IRL applies the principle of maximum entropy to select the probability distribution over trajectories that matches the expert's empirical feature expectations while making no other assumptions (i.e., being as random as possible otherwise).

Under the MaxEnt framework, the probability of an agent choosing a specific trajectory $\xi$ is exponentially proportional to the total accumulated reward of that trajectory:
\begin{equation}
    P(\xi \mid w) = \frac{1}{Z(w)} \exp \left( \sum_{(s,a) \in \xi} w^\top \phi(s, a) \right) = \frac{1}{Z(w)} e^{R_w(\xi)}
\end{equation}
where $Z(w)$ is the partition function, representing the integral (or sum) over all possible trajectories originating from the start state:
\begin{equation}
    Z(w) = \sum_{\xi' \in \Xi} e^{R_w(\xi')}
\end{equation}

The reward weights $w$ are then found by maximizing the log-likelihood of the demonstrated trajectories $\mathcal{D}$:
\begin{equation}
    \mathcal{L}(w) = \sum_{i=1}^{|\mathcal{D}|} \log P(\xi_i \mid w) = \sum_{i=1}^{|\mathcal{D}|} \left( w^\top \phi(\xi_i) - \log Z(w) \right)
\end{equation}
The gradient of this log-likelihood neatly reduces to the difference between the empirical feature counts of the demonstrations and the expected feature counts under the current weight vector $w$, allowing for efficient optimization via gradient ascent.

%% file: Sections/Approach.tex
\section{Multi-Objective Constraint Inference (MOCI)}
\label{sec:methods}
In this section, we present our approach for jointly learning shared constraints and individual preferences. We also provide a theoretical analysis of the computational complexity of the proposed algorithm, highlighting its scalability with respect to the number of states, actions, and demonstrations.
\subsection{Approach}
Let $\mathcal{D} = \{\xi_1, \dots, \xi_N\}$ be a dataset of demonstrated trajectories of maximum length $H$. We assume a \textit{ Constraint Multi-Objective Markov Decision Process} (CMOMDP) $\langle\mathcal{S},\mathcal{A},\mathcal{T},\gamma,\boldsymbol{R},C\rangle$ and the existence of $K$ latent expert types (clusters), where each type $k \in \{1, \dots, K\}$ is characterized by a specific preference weight vector $w_k$ such that $R_k=w_k^\top \boldsymbol{R}$, and a prior probability $\pi_k = P(k)$. Crucially, the demonstrations are not labelled by their associated expert type $k \in K$ as inferring expert preferences is a key goal of this work.

Although experts differ in their preferences, all agents share the same state-action feature function $\phi$, and operate subject to a shared set of hard physical constraints, denoted by $C$. Following the Maximum Entropy Inverse Reinforcement Learning (MaxEnt IRL) framework by~\citet{ziebart2008maximum}, the probability of observing a specific trajectory $\xi$, given that it was generated by an expert of type $k$ subject to constraints $C$, is defined as:
\begin{equation}
    P(\xi \mid C, w_k) = \frac{1}{Z(C, w_k)} e^{R_{w_k}(\xi)} \mathbb{I}^C(\xi)
\end{equation}
where, $ 
R_{w_k}(\xi) = \sum_{(s,a) \in \xi} w_k^\top \phi(s, a)
$
is the cumulative reward of the trajectory under preference $w_k$. $Z(C, w_k)$ is the partition function over all feasible paths in the constrained Markov Decision Process (MDP). The detailed analysis and proof are given in the appendix-\ref{appendix-Proof}. $\mathbb{I}^C(\xi)$ is an indicator function that equals $1$ if the trajectory $\xi$ does not violate any constraint in $C$ and $0$ otherwise as presented in equation \eqref{equ1}.
Marginalizing over the latent assignment of demonstrations to expert types, the likelihood of a single demonstration is:
\begin{equation}
    P(\xi \mid C, \{w_k\}, \{\pi_k\}) = \sum_{k=1}^K \pi_k \frac{e^{R_{w_k}(\xi)}}{Z(C, w_k)} \mathbb{I}^C(\xi)
\end{equation}

Our objective is to jointly infer the shared constraints $C$, the latent reward weights $\{w_k\}$, and the priors $\{\pi_k\}$ by maximizing the joint log-likelihood of the dataset $\mathcal{D}$:
\begin{equation}
    \mathcal{L}(C, \{w_k\}, \{\pi_k\}) = \sum_{i=1}^{|\mathcal{D}|} \log \left( \sum_{k=1}^K \pi_k \frac{e^{R_{w_k}(\xi_i)}}{Z(C, w_k)} \mathbb{I}^C(\xi_i) \right)
    \label{eq:joint_log_likelihood}
\end{equation}

Because the assignments of the demonstrations to agent types or clusters are unobserved, directly optimizing this joint log-likelihood is intractable. We therefore optimize equation \eqref{eq:joint_log_likelihood} using an Expectation-Maximization (EM) approach \cite{hamidi2018privacy}, which alternates between estimating the posterior probability of cluster assignments and updating the model parameters alongside the constraint set.  

In the Expectation step (E-step), we fix the current constraints $C$, weights $\{w_k\}$, and priors $\{\pi_k\}$, and compute the responsibility $\gamma_{i,k}$, which represents the posterior probability that trajectory $\xi_i$ was generated by expert $k$:
\begin{equation}
    \gamma_{i,k} = \frac{\pi_k P(\xi_i \mid C, w_k)}{\sum_{j=1}^K \pi_j P(\xi_i \mid C, w_j)}
\end{equation}
In the Maximization step (M-step), we update the parameters to maximize the expected log-likelihood over the entire dataset. The cluster priors are updated as the empirical mean of the responsibilities, such that
\begin{equation}
    \pi_k = \frac{1}{|\mathcal{D}|} \sum_{i=1}^{|\mathcal{D}|} \gamma_{i,k}
\end{equation}
To update the reward weights $w_k$ for each cluster, we perform gradient ascent where the gradient for trajectory $i$ and cluster $k$ is weighted by the responsibility $\gamma_{i,k}$, ensuring that the weights adapt to the trajectories probabilistically assigned to that cluster:
\begin{equation}
    \nabla_{w_k} \mathcal{L} = \sum_{i=1}^{|\mathcal{D}|} \gamma_{i,k} \left( \phi(\xi_i) - \mathbb{E}_{P(\xi \mid C, w_k)}[\phi(\xi)] \right)
\end{equation}
where $\mathbb{E}_{P(\xi \mid C, w_k)}[\phi(\xi)]$ is the expected feature count under the current constraints and cluster weights. 
$w_k \leftarrow w_k + \alpha \nabla_{w_k} \mathcal{L}$

Finally, to update the shared constraints $C$, we execute a greedy search over the set of candidate constraints (states not visited by any expert in $\mathcal{D}$). For a candidate constraint $c$, we define its score as the joint log-likelihood evaluated with the augmented constraint set;
\begin{equation}
    \text{Score}(c) = \mathcal{L}(C \cup \{c\}, \{w_k\}, \{\pi_k\})
\end{equation} 
The algorithm iteratively adds the candidate $\hat{c} = \arg\max_{c} \text{Score}(c)$ to $C$.
\\
This greedy addition terminates when the reduction in Kullback-Leibler (KL) divergence falls below a specified threshold $d_{DKL}$, which is mathematically equivalent to stopping when the increase in log-likelihood satisfies;
\begin{equation}
    \mathcal{L}(C \cup \{\hat{c}\}, \cdot) - \mathcal{L}(C, \cdot) \le d_{DKL}
\end{equation} 
This thresholding mechanism serves as a regularizer, preventing the algorithm from overfitting to noise by rejecting constraints that yield only marginal improvements to the model's likelihood. 

The algorithm for MOCI is presented in appendix -\ref{Sec:em_mlci_algo}. The goal of the algorithm is to employs an Expectation-Maximization (EM)  approach to infer shared constraints and multi-objective preferences.

%% file: Sections/ComputationCost.tex
\subsection{Computational Complexity}
The MOCI algorithm calculates the likelihood of each trajectory in $\mathcal{D}$ demonstrations belonging to each group $k \in K$. It computes partition function $Z$, which iterates through the maximum horizon (trajectory length) $H$, checking all actions $|\mathcal{A}|$ for all valid states $|\mathcal{S}|$. Which yields complexity: 
\begin{equation}
    \mathcal{O}_E (K \cdot H \cdot (|\mathcal{S}| \cdot |\mathcal{A}| + |\mathcal{D}|))
\end{equation} 
For each cluster $K$, the algorithm performs $I_{IRL}$ gradient descent steps. In each step, it recomputes $Z$ to find the expected feature counts and compares them to the empirical features of the demonstrations. It can be shown as: 
\begin{equation}
\mathcal{O}_M (K \cdot I_{IRL} \cdot H \cdot (|\mathcal{S}| \cdot |\mathcal{A}| + |\mathcal{D}|))
\end{equation}
The algorithm also tests unvisited candidate states to see if adding them to the constraint set $\mathcal{C}$ reduces the log-likelihood by less than threshold ($d_{DKL}$). Let $\mathcal{C}_{added}$ be the number of constraints successfully inferred in a single step. For each added constraint, the algorithm tests a subset of candidates ($M_{test}$).
\begin{equation}
    \mathcal{O}_{\mathcal{C}}(\mathcal{C}_{added} \cdot M_{test} \cdot K \cdot |\mathcal{S}| \cdot |\mathcal{A}| \cdot H)
\end{equation}
To find the total computational cost, we sum the complexities of all the steps (Expectation step $\mathcal{O}_E$, Weights $\mathcal{O}_M$, and Constraints $\mathcal{O}_{\mathcal{C}}$), and multiply by the total number of iterations ($I_{Iter}$). 
\begin{equation}
    \text{Cost} = \mathcal{O} \Big( I_{Iter} \cdot K \cdot |\mathcal{S}| \cdot |\mathcal{A}| \cdot H \cdot (1 + I_{IRL} + \mathcal{C}_{added} \cdot M_{test}) \Big)
\end{equation}
To simplify complexity equation, we isolate the terms that grow the fastest as the problem scales toward infinity. In Inverse Reinforcement Learning, the hyper-parameters ($I_{Iter}, K, I_{IRL}, M_{test}, \mathcal{C}_{added}$) can be treated as constants. Furthermore, the size of the state-action space multiplied by the horizon ($|\mathcal{S}| \cdot |\mathcal{A}| \cdot H$) will always asymptotically dominate the number of sampled demonstrations ($|\mathcal{D}|$).
\\
By dropping the constants and non-dominant terms, the total theoretical computational cost reduces to:
\begin{equation}
    \text{Cost} = \mathcal{O}(|\mathcal{S}| \cdot |\mathcal{A}| \cdot H)
\end{equation}
The cost of MOCI is highly dependent on the total number of states $|\mathcal{S}|$, the total number of actions per state $|\mathcal{A}|$, and the horizon (maximum length of a trajectory) $H$. For a specific environment, e.g., Gridworld, the total number of states $|\mathcal{S}| = N^2$, hence the MOCI algorithm scales quadratically with the grid dimensions. 

%% file: Sections/ExperimentAndResults.tex
% \begin{figure}[t]
%     \centering
%     \includegraphics[height=5cm, width=10cm]{Diagrams/True_GridWorld_Env_and_Demos.png}
%     \caption{Demonstration of the MOCI algorithm in a 6x6 Gridworld. Ground-truth environment with water hard constraints (blue) and expert trajectories for a Grass-Lover (lime) and Rock-Lover (orange).}
%     \label{fig:True_GridWorld_Env}
% \end{figure}
%%%%%%%%%%%
% \begin{figure}
%     \centering
%     \includegraphics[height=5cm, width=10cm]{Diagrams/MOCI-inferred-constraints.png}
%     \caption{MOCI-inferred constraints (red hatched), showing successful recovery of true constraints alongside false positives in unvisited states.}
%     \label{fig:MOCI-inferred constraints}
% \end{figure}
%%%%%%%%%%%%%%%%%%%%%%%%%%%%%%%%%%%%%%%%%%
\begin{figure}[t] % [t] places the combined figure at the top of the page
    \centering
    % --- First Figure (Left) ---
    \begin{minipage}{0.48\textwidth}
        \centering
        % width=\linewidth automatically scales the image to fit the minipage
        \includegraphics[width=\linewidth]{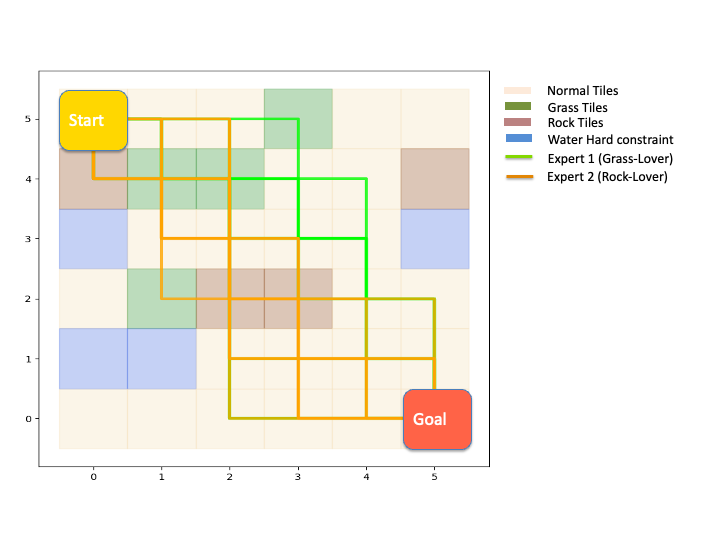}
        \caption{Demonstration of the MOCI algorithm in a 6x6 Gridworld. Ground-truth environment with water hard constraints (blue) and expert trajectories for a Grass-Lover (lime) and Rock-Lover (orange).}
        \label{fig:True_GridWorld_Env}
    \end{minipage}\hfill % \hfill adds flexible spacing between the two images
    % --- Second Figure (Right) ---
    \begin{minipage}{0.48\textwidth}
        \centering
        \includegraphics[width=\linewidth]{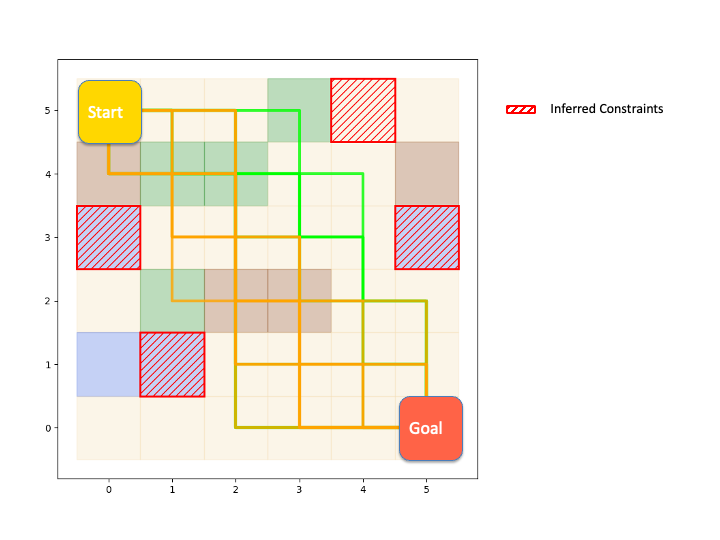}
        \caption{MOCI-inferred constraints (red hatched), showing successful recovery of true constraints alongside false positives in unvisited states.}
        \label{fig:MOCI-inferred constraints}
    \end{minipage}
\end{figure}
%%%%%%%%%%%%%%%
\section{Experiments: Heterogeneous Gridworld }
\label{sec:experiments}
To validate the Multi-Objective Constraint Inference (MOCI) framework, we designed a simulated multi-feature Gridworld environment. This environment serves as a controlled testbed for evaluating the MOCI algorithm.
%%%%%%%%%%%%%%%%%%%%%%%%%%%%%%%%
\subsection{Environment Setup}
The environment is modeled as a deterministic Markov Decision Process (MDP) on a $N \times N$ grid as shown in figure-\ref{fig:True_GridWorld_Env}. The state space $\mathcal{S}$ consists of grid cells, and the action space $\mathcal{A}$ allows for standard movement: $\{\textit{Up, Down, Left, Right, Stay}\}$. The environment features a "Start" state in the top-left corner and a "Goal" state in the bottom-right corner. 
\\
Each state $s \in \mathcal{S}$ is assigned a categorical feature vector $\phi(s) \in \{0, 1\}^4$ that represents four different types of terrain: grasslands (green), Rocks(brown), Water (blue) and Normal (light beige) tiles. Exactly one feature is active for any given state. The true underlying environment contains a set of impassable obstacles (blue water tiles in this case) that represent our set of ground-truth hard constraints $C^*$. Transitions into states $s \in C^*$ are strictly prohibited in the environment dynamics.
\\
To simulate a heterogeneous population, we defined two distinct expert types with divergent reward preferences, parameterized by their ground-truth weight vectors $w_k^*$:
\begin{enumerate}
    \item \textbf{Expert 1 (Grass-Lover):} Highly prefers traveling on the Grass tiles
    \item \textbf{Expert 2 (Rock-Lover):} Highly prefers traveling on the Rock tiles
\end{enumerate}
The agent's objectives are to reach the goal state by taking the shortest path while navigating their preferred terrain. Both experts share the same terminal goal state and are fundamentally restricted by the Water tiles ($C^*$). 
\\
To generate the demonstration dataset $\mathcal{D}$, we computed the optimal Maximum Entropy policy $\pi_{k}^*$ for each expert using Soft Value Iteration based on their respective reward weights $w_k^*$. We then sampled $|\mathcal{D}_1| = 10$ trajectories from Expert 1 and $|\mathcal{D}_2| = 10$ trajectories from Expert 2. The final dataset provided to the inference algorithms was an unlabeled, randomly shuffled mixture of these 20 trajectories, $\mathcal{D} = \mathcal{D}_1 \cup \mathcal{D}_2$. In Figure-\ref{fig:True_GridWorld_Env}, the trajectory of Expert 1 (Grass-Lover) is shown in lime, while the trajectory of Expert 2 (Rock-Lover) is shown in orange.
%%%%%%%
\subsection{Results}
\begin{figure*}[t]
\centering
\includegraphics[height=5cm, width=14cm]{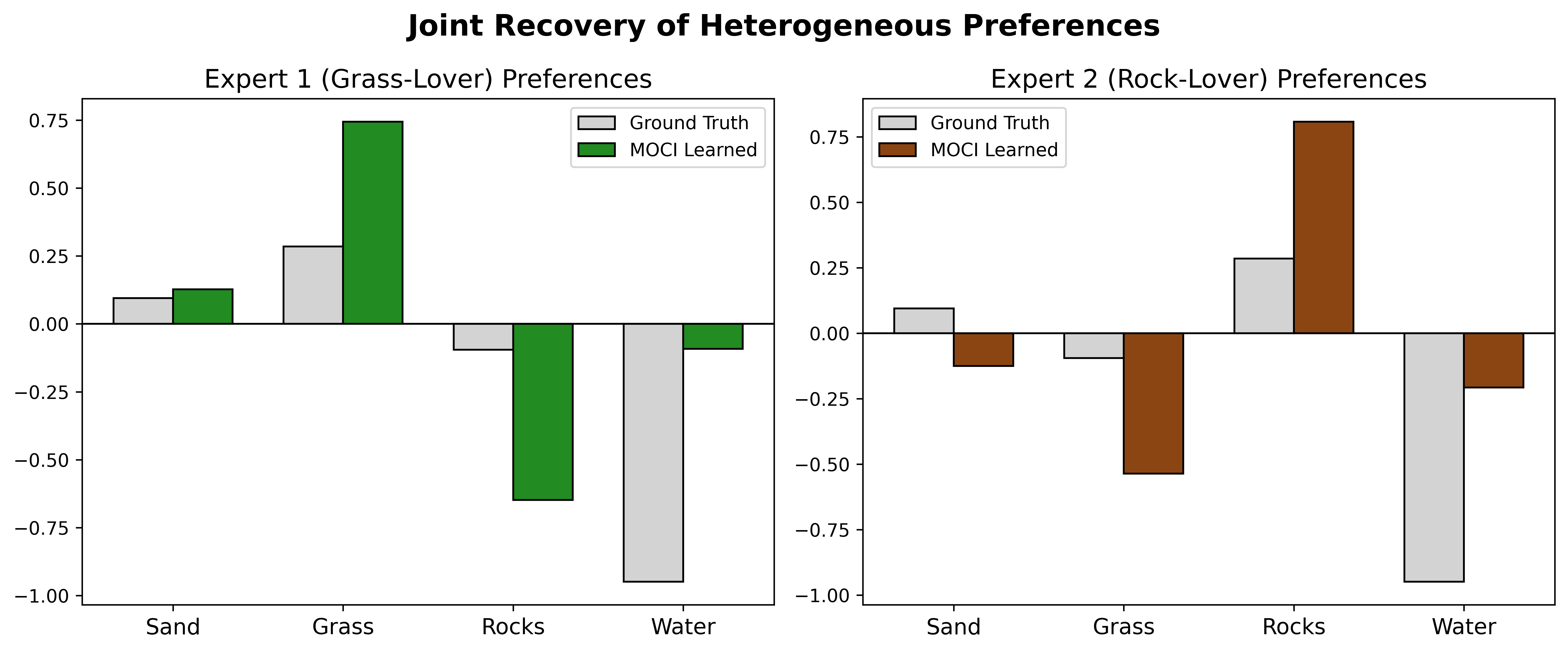}
\caption{Joint recovery of heterogeneous preferences using the MOCI algorithm. The chart compares the ground-truth reward weights with the learned weights across four terrain features for two distinct expert profiles (a Grass-Lover and a Rock-Lover).}
\label{fig:prefernce_learning}
\vspace{-1.5em}
\end{figure*}

All results reported in this paper were obtained by running experiments on a \textit{MacBook M3} using \textit{Python 3.11.5}. The figure-\ref{fig:MOCI-inferred constraints} visualizes the constraints inferred by the MOCI algorithm after observing the expert trajectories shown in figure-\ref{fig:True_GridWorld_Env}.
The algorithm successfully places "Inferred Constraints" (marked by red hatched boxes) over the blue water tiles. By observing that neither the grass-lover nor the rock-lover ever stepped on these tiles, MOCI correctly deduced they are forbidden.
\\
The red hatched boxes also appear on several non-water tiles (e.g., a normal tile on the top right). This occurs because these specific location was never visited by either expert. The algorithm conservatively (and incorrectly) inferred that these unvisited states must also be hard constraints.
It shows that while MOCI can successfully learn heterogeneous preferences and recover ground-truth constraints (the water), limited dataset coverage can lead the algorithm to hallucinate constraints in regions of the state space that were simply ignored by the expert's specific goals.
\\
The figure-\ref{fig:prefernce_learning} illustrates the MOCI algorithm's capability to untangle and learn the distinct underlying reward functions of different types of experts from a shared environment. The visualization is split into two side-by-side bar charts, each representing a specific expert profile evaluated across four terrain features: Sand, Grass, Rocks, and Water. This left-hand chart compares the original Ground Truth weights (light gray bars) against the MOCI Learned weights (green bars).
The algorithm accurately captures the expert's defining trait: a strong positive preference for Grass.
It also correctly identifies the negative polarity for Rocks and Water, meaning the algorithm learned that the expert avoids these terrains. The right-hand chart contrasts the Ground Truth weights (light gray bars) with the MOCI Learned weights (brown bars) for the second expert.
Here, the algorithm successfully identifies the opposite behavior, showing a strong positive preference for Rocks.
It correctly assigns negative weights to Grass and Water, capturing the expert's specific aversions.
\\
\textbf{Tuning Threshold $d_{DKL}$:} The adjustment of the threshold value is arguably the most critical step in the implementation of MOCI. The threshold ($d_{DKL}$ dictates how much the likelihood of a trajectory must drop before the algorithm considers a state to be a hard constraint. If it is too high, the algorithm misses the real walls (false negatives). If it is too low, the algorithm looks at a perfectly navigable tile that the experts simply disliked and declares it a physical wall (a false positive).
We have presented some of the factors on which the performance of MOCI is really dependent in Appendix-\ref{sec:tuning_ddkl}. 
\\
To ensure that the threshold remains invariant with the size of the dataset $|\mathcal{D}|$, we normalize the joint log-likelihood to represent the average information gained per trajectory:
\begin{equation}
\label{equ:normalized_log_liklyhod}
    L_{avg} = \frac{1}{|\mathcal{D}|} \sum_{i=1}^{|\mathcal{D}|} \log P(\xi_i | w_k, C)
\end{equation}
By evaluating the normalized change, $\Delta L_{avg}$, the divergence threshold $d_{DKL}$ can be robustly set to a constant value across varying expert population sizes. Established literature suggests an effective range of $0.01 \leq d_{DKL} \leq 2$ for this parameter.
%%%%%%%%%%%%%%%%%%%%%%%%%%%%
%%%%%%%%%%%%%%%%%%%%%%%%%%%%%%%%
%%%%%%%%%%%%%%%%%%%%%%%%%%%%%%%%%%%%%%%
\begin{figure}[b] % [t] forces the block to the top of the page
    \centering
    % --- First Figure ---
    \begin{minipage}{0.48\textwidth}
        \centering
        \includegraphics[width=\linewidth]{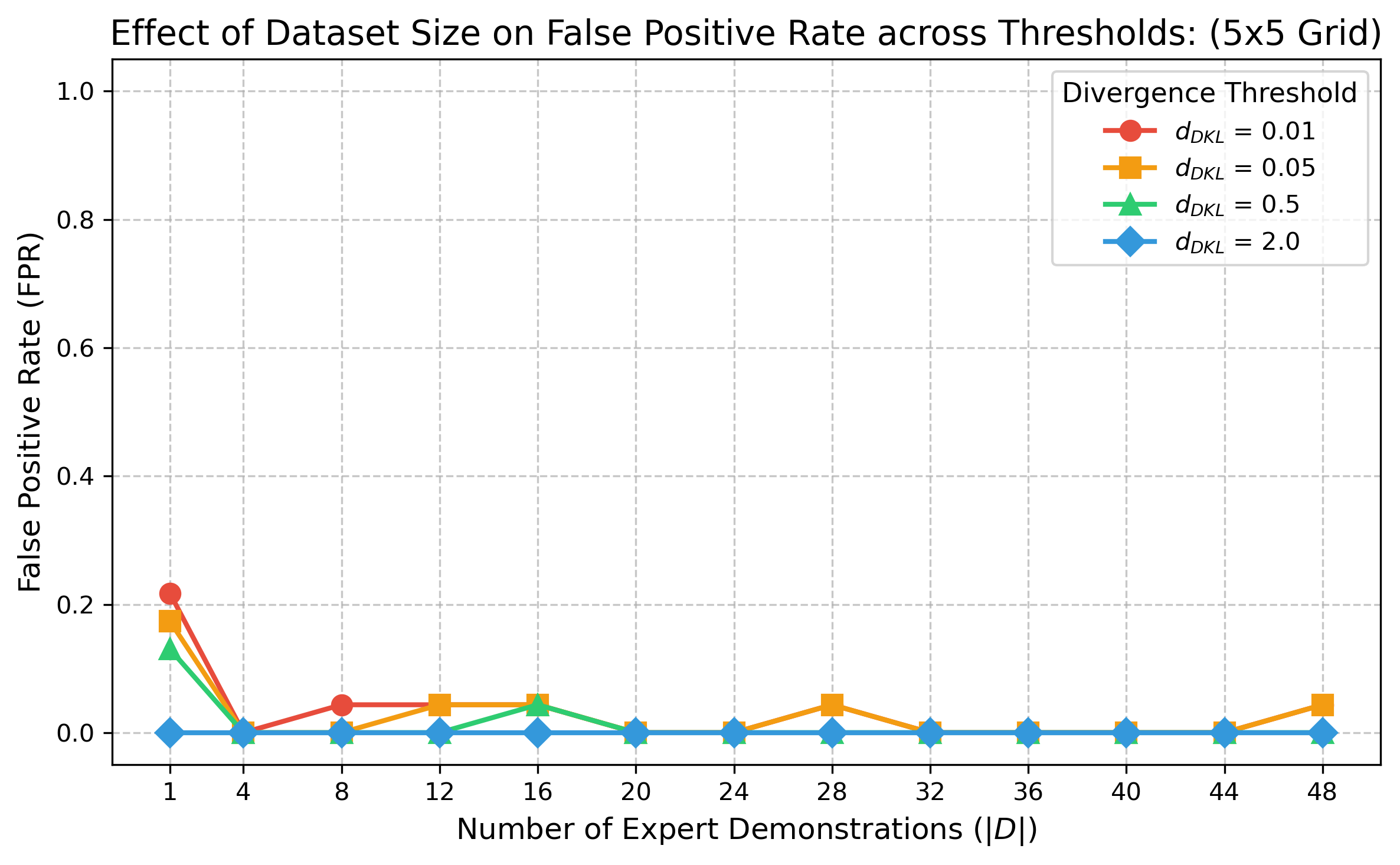}
        \caption{Effect of the number of expert demonstrations ($|\mathcal{D}|$) on the False Positive Rate (FPR) during constraint inference evaluated across four thresholds ($d_{DKL}$) in a 5x5 Gridworld.}
        \label{fig:Threshhold_Plot}
    \end{minipage}\hfill % \hfill creates a flexible horizontal space between the two minipages
    % --- Second Figure ---
    \begin{minipage}{0.48\textwidth}
        \centering
        \includegraphics[width=\linewidth]{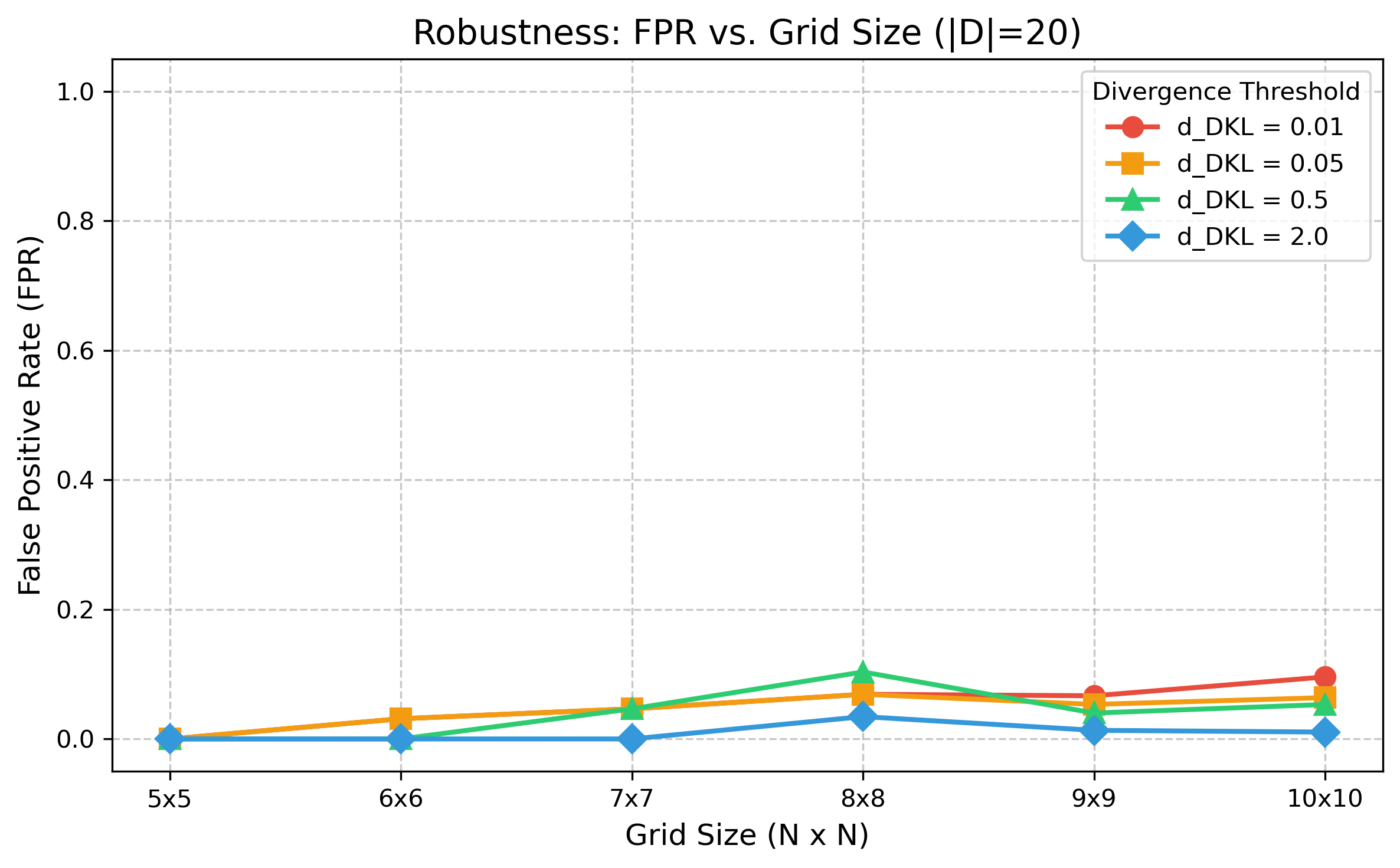}
        \caption{Robustness of the algorithm's False Positive Rate (FPR) relative to environmental scaling (grid sizes from 5x5 to 10x10) given a fixed dataset of $|D| = 20$ expert demonstrations.}
        \label{fig:FPR_vs_GridSize}
    \end{minipage}
\end{figure}
%%%%%%%%%%%%%%%%%%%%%%%%%%%%%%
% \begin{figure}
%     \centering
%     \includegraphics[width=.48\linewidth]{Diagrams/Runtime_vs_GridSize_Dynamic_Horizons.png}
%     \caption{Scalability analysis of run-time versus grid size with dynamically scaling trajectory lengths. The execution time exhibits a steepening curve, effectively validating the theoretical complexity $\mathcal{O}(|S| \cdot |A| \cdot H)$.}
%     \label{fig:Runtime_vs_GridSize}
% \end{figure}
%%%%%%%%%%%%%%%%%%%%%%%%%%%%%%

\begin{figure}[t] % [t] pushes the whole block to the top of the page
    \centering
    
    % --- Left Side: The Figure ---
    \begin{minipage}[c]{0.48\textwidth}
        \centering
        \includegraphics[width=\linewidth]{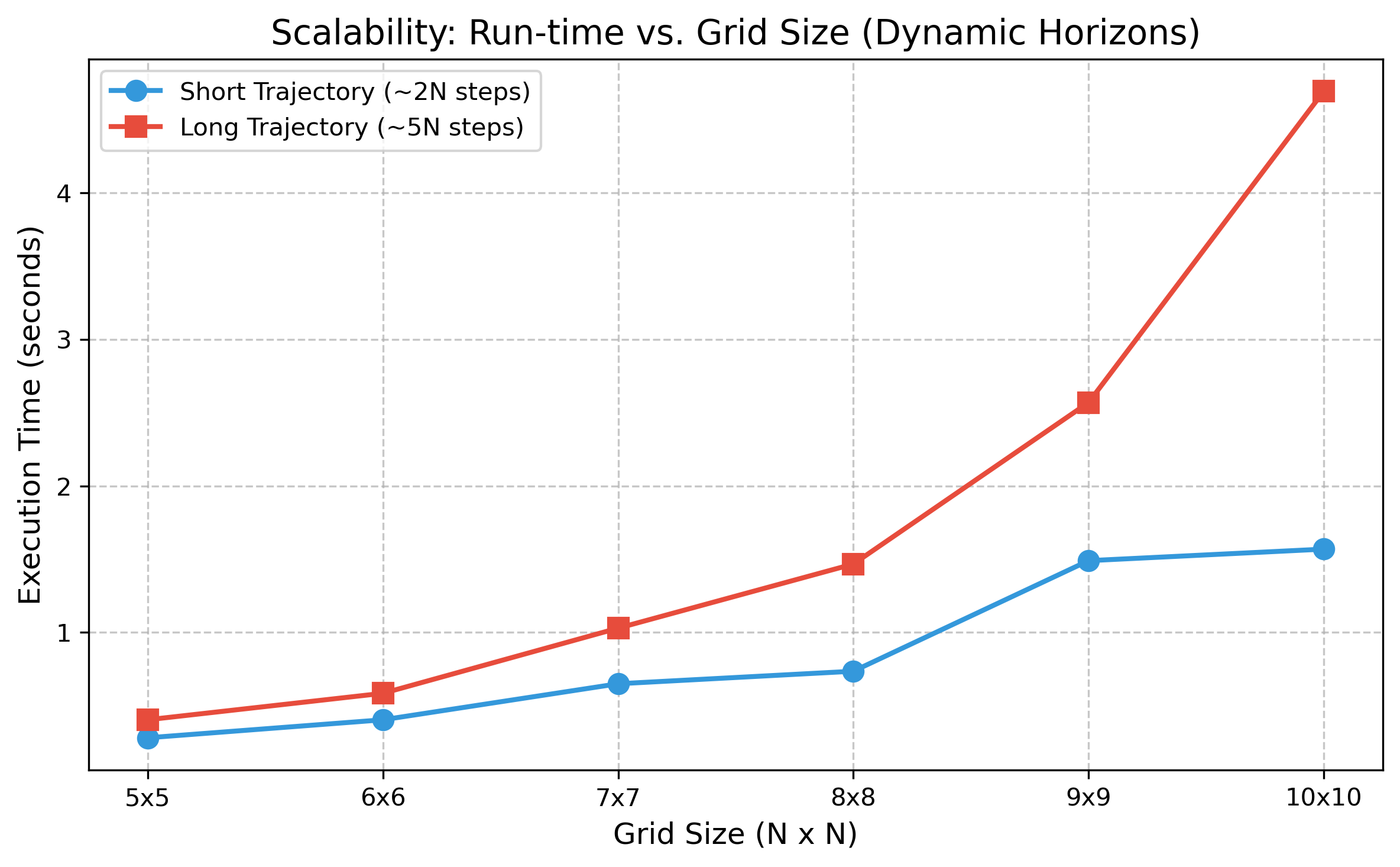}
        \captionof{figure}{Scalability analysis of run-time versus grid size with dynamically scaling trajectory lengths. The execution time exhibits a steepening curve, effectively validating the theoretical complexity $\mathcal{O}(|\mathcal{S}| \cdot |\mathcal{A}| \cdot H)$.}
        \label{fig:Runtime_vs_GridSize}
    \end{minipage}\hfill % \hfill creates flexible spacing between the two minipages
    %
    % --- Right Side: The Table ---
    \begin{minipage}[c]{0.48\textwidth}
        \centering
        % Table captions traditionally go ABOVE the table
        \captionof{table}{Performance Comparison with existing techniques}
        \label{Comparison_table}
        \vspace{2mm} % Adds a tiny gap between caption and table
        
        % \resizebox scales the wide table to fit inside the 0.48\textwidth minipage
        \resizebox{\linewidth}{!}{
            \renewcommand{\arraystretch}{1.5} 
            \setlength{\tabcolsep}{6pt} % Reduced column separation
            \begin{tabular}{l c c c c}
            \rowcolor[HTML]{EFEFEF} 
            \toprule
            \textbf{Techniques} & \textbf{CMSE} & \textbf{Run-time} & \textbf{Trajectory Types} & \textbf{Preference Learning} \\ 
            \midrule
            MLCI \cite{mcpherson2021maximum}  & 0.25 & 0.2927 & Homogeneous & No \\
            ICRL \cite{malik2021inverse}  & 0.36    & 8.90 & Homogeneous & No \\
            MOCI (our) &  0.027  & 0.6917 & Heterogeneous & Yes \\
            \bottomrule
            \end{tabular}
        }
    \end{minipage}
    
\end{figure}

Figure-\ref{fig:Threshhold_Plot} illustrates the False Positive Rate (FPR) with respect to the four different thresholds ($d_{DKL} \in \{0.01, 0.05, 0.5, 2.0\}$). The performance of the algorithm is evaluated on different number of Expert Demonstrations ($|\mathcal{D}|$).
Despite the different threshold values, the FPR does not escalate as the number of expert demonstrations increases. Instead, the metric remains highly stable and shows a trend toward convergence. When the algorithm receives very few demonstrations (e.g. $|\mathcal{D}| = 1$), there is a higher degree of uncertainty. This lack of data causes the algorithm to over-infer constraints, resulting in a higher FPR.  As the number of expert demonstrations increases, the FPR drops and converges. The highest threshold ($d_{DKL} = 2.0$, represented by the blue line) consistently yields the lowest False Positive Rate for this particular gird size. This proves that the algorithm successfully normalizes the log-likelihood against the dataset size, preventing metric inflation as more data are introduced.
\\
\textbf{Robustness:} Figure~\ref{fig:FPR_vs_GridSize} presents an evaluation of the algorithm’s robustness in terms of its False Positive Rate (FPR) as the environment scales (grid sizes), while keeping the amount of expert data fixed at $|\mathcal{D}| = 20$ demonstrations.
\\
As the grid scales from $5\times5$ to $10\times10$, the total number of states increases quadratically (from 25 to 100). Consequently, a fixed dataset of 20 trajectories covers a progressively smaller percentage of the environment. Because the expert's coverage becomes sparser in larger grids, the algorithm's uncertainty increases slightly, leading to a minor upward trend in the FPR. Additionally, despite the state space quadrupling in size, the FPR remains remarkably low.
\\
\textbf{Scalability effect on runtime:} Figure-\ref{fig:Runtime_vs_GridSize} illustrates the execution time (in seconds) of the MOCI algorithm as the complexity of the environment increases from a $5\times5$ grid to a $10\times10$ grid. The blue line (short trajectory) represents an optimal direct path to the goal, where the horizon scales roughly twice the dimension of the grid ($H \approx 2N$).
The red line (long trajectory) represents a sub-optimal or highly exploratory path, where the horizon could scale at five times the grid dimension ($H \approx 5N$). For optimal paths ($~2N$), the computational cost increases at a manageable rate. 
\\
In a $5\times5$ grid, the execution takes roughly 0.3 seconds, and it smoothly increases to just under 1.6 seconds for a $10\times10$ grid. For exploratory paths ($~5N$), the algorithm requires significantly more computation. The execution time curves increase much more sharply, starting around 0.4 seconds for a $5\times5$ grid but rising to nearly 4.7 seconds for a $10\times10$ grid. This graph perfectly validates the theoretical computational cost previously defined as $\mathcal{O}(N^2 \cdot |\mathcal{A}| \cdot H)$.
\\
In summary, these findings validate MOCI as a uniquely robust, scalable, and computationally viable framework for extracting shared constraints from complex, heterogeneous expert behaviors.
%%%%%%%%
% \begin{table*}[t]
% \centering
% \caption{Performance Comparison with existing techniques}
% \renewcommand{\arraystretch}{1.5} 
% \setlength{\tabcolsep}{10pt} % Slightly reduced to fit more columns
% \begin{tabular}{l c c c c}
% \rowcolor[HTML]{EFEFEF} 
% \toprule
% \textbf{Techniques} & \textbf{CMSE} & \textbf{Run-time} & \textbf{Trajectory Types} & \textbf{Preference Learning} \\ 
% \midrule
% MLCI \cite{mcpherson2021maximum}  & 0.25 & 0.2927 & Homogeneous & No \\
% ICRL \cite{malik2021inverse}  & 0.36    & 8.90 & Homogeneous & No \\
% MOCI (our) &  0.027  & 0.6917 & Heterogeneous & Yes \\
% \bottomrule
% \end{tabular}
% \label{Comparison_table}
% \end{table*}
%%%%%%%%%%%%%%%%%%%%%%%%%%%%

%% file: Sections/Comparison_with_Baseline.tex
\section{Comparison with existing techniques} \label{Sec:comparison}
To rigorously evaluate the efficacy of the proposed Multi-Objective Constraint Inference (MOCI) framework, it is imperative to benchmark its performance against established state-of-the-art techniques.
Table-\ref{Comparison_table} presents the performance comparison with existing techniques. It compares the MOCI techniques with the two existing baselines, Maximum Likelihood Constraint Inference (MLCI) \cite{mcpherson2021maximum} and Inverse Constrained Reinforcement Learning (ICRL) \cite{malik2021inverse} across four evaluation metrics.

\textbf{Constraint mean squared error (CMSE)} is computed as the mean squared error between the true constraint function and the recovered constraint function.
\\
\textbf{Run-time} represents the execution or computational time required by the technique, including training and inferring the constraints.
\\
\textbf{Trajectory Types} indicates whether the model handles "Homogeneous" or "Heterogeneous" data.
\\
\textbf{Preference Learning} is a boolean metric indicating whether the technique utilizes preference learning ("Yes" or "No").
\\
We measured these performace matrices over a $5\times5$ gridworld and results are summarized in Table-\ref{Comparison_table}. Our proposed MOCI framework demonstrates superior predictive accuracy, achieving a Mean Squared Error (MSE) of 0.027, which represents a substantial improvement over the baseline techniques MLCI (0.25) and ICRL (0.36). Although MLCI remains the most computationally efficient model with a run-time of 0.2927, MOCI maintains a highly competitive run-time of 0.6917.
This is because in MLCI, there are no explicit weight updates during the constraint inference loop. The nominal MDP has the expert's reward function completely known \textit{a priori}. The only thing that changes is the set of forbidden states. On the other hand, MOCI maintains a list of weights and it performs gradient descent to actively update the reward parameters for each cluster based on the responsibilities calculated in the each step. ICRL has a comparatively longer run-time (8.90), mainly due to its neural network–based soft parameterization of the indicator set over constrained trajectories. 
\\
Furthermore, unlike MLCI and ICRL, which are restricted to homogeneous trajectory types and lack preference learning capabilities, MOCI is uniquely able to process heterogeneous trajectories while explicitly incorporating preference learning. These advantages highlight MOCI's enhanced flexibility and robustness without sacrificing computational efficiency. 

%% file: Sections/Conclusion.tex
\section{Conclusion} \label{Sec:conclusion}
In this study, we presented MOCI, a novel Multi-Objective Constraint Inference framework that successfully addresses the limitations of current constraint learning techniques. By leveraging preference learning, MOCI is uniquely process heterogeneous trajectory data, allowing it to accurately capture diverse expert behaviors and nuanced multi-objective trade-offs. Our comparative analysis reveals that MOCI achieves superior predictive accuracy, reducing the Mean Squared Error to 0.027. This is a substantial improvement over established baselines. Crucially, MOCI achieves this high accuracy while bypassing the severe computational overhead associated with iterative reinforcement learning loops, resulting in highly competitive execution times. These findings highlight MOCI's potential as a scalable and robust tool for safe reinforcement learning in complex environments where expert data is varied and multifaceted.

\textbf{Limitations:}
\begin{enumerate}[noitemsep, topsep=0pt]
    \item MOCI currently requires the hyperparameter $K$ (the number of distinct expert types ) to be defined a priori. In real-world, unannotated datasets, the exact number of heterogeneous behaviors is rarely known in advance. If $K$ is set incorrectly, MOCI may either force distinct behaviors into a single distorted reward function or split a single behavior into redundant clusters. 
    \item The MOCI focuses strictly on hard constraints (states that are absolutely forbidden). It does not natively support the inference of soft constraints (e.g., a speed limit that an expert might occasionally violate but generally avoids) or probabilistic chance constraints.
\end{enumerate}
\textbf{Future work:}
\begin{enumerate}[noitemsep, topsep=0pt]
    \item To eliminate the need for a predefined number of expert types, the EM clustering step could be replaced with Bayesian non-parametric methods, such as a Dirichlet Process Mixture Model (DPMM) \cite{niekum2011clustering}. This would allow the algorithm to automatically deduce the optimal number of distinct expert profiles ($K$) directly from the complexity and diversity of the demonstration data.
    \item This work can be extended to real-world healthcare applications. In cancer treatment, patients are typically subject to shared clinical constraints, such as treatment guidelines, toxicity limits, and eligibility criteria for specific therapies (e.g., chemotherapy, radiotherapy). However, individual preferences can vary substantially. Some patients may prioritize prolonging survival, even at the cost of aggressive treatments and severe side effects, while others may prefer to preserve quality of life and opt for less intensive treatments. Furthermore, treatment effects are highly heterogeneous \cite{ye2025deep}, as patients may respond differently to the same therapy due to clinical, biological, and personal factors. By leveraging MOCI to learn shared constraints, we aim to optimize treatment strategies that align with individual patient preferences while adhering common constraints and safety boundaries.
    % \item This work can be extended to real-world healthcare settings. In cancer treatment, patients are subject to shared clinical constraints, such as guidelines, toxicity limits, and therapy eligibility. However, patient specific objective and preferences vary significantly. Some patients prioritize survival and others focus on quality of life. Treatment responses are also highly heterogeneous \cite{ye2025deep}. Using MOCI to learn shared constraints, we aim to optimize treatment strategies that align with individual patient preferences while adhering to safety boundaries(constraints). 
\end{enumerate}
 

%% file: Sections/Apendics.tex
 \section*{\centering Appendices}

\section{Notations and description}
Table-\ref{table:Notations} summarizes the notation used in this paper along with their descriptions. 
\begin{table}[htbp]
    \centering
    \caption{Notations and Descriptions for Multi-Objective Constraint Inference (MOCI)}
    \label{table:Notations}
    \begin{tabularx}{\textwidth}{lX}
        \toprule
        \textbf{Notation} & \textbf{Description} \\
        \midrule
        $\mathcal{M}$ & Markov Decision Process (MDP)  \\
        $\mathcal{S}$ & State space  \\
        $\mathcal{A}$ & Action space  \\
        $\mathcal{T}(s'|s,a)$ & Transition probability distribution  \\
        $\gamma$ & Discount factor  \\
        $R(s,a)$ & Reward function  \\
        $\pi(a|s)$ & Policy  \\
        $C$ & Set of forbidden or unsafe states (constraints)  \\
        $\mathcal{M}_C$ & Constrained Markov Decision Process (CMDP)  \\
        $\xi$ & Trajectory of state-action pairs  \\
        $H$ & Length of a trajectory  \\
        $\mathcal{I}_C(\xi)$ & Binary indicator function for constraint violation  \\
        $\phi(s, a)$ & Vector of state-action features  \\
        $w$ & Preference weight vector for scalarization  \\
        $R_w(s, a)$ & Scalarized reward function for preference $w$  \\
        $k \in \{1,...,K\}$ & Latent expert types or clusters  \\
        $P(\xi|w)$ & Probability of a trajectory under Maximum Entropy IRL  \\
        $Z(w)$ & Partition function (sum over all possible trajectories)  \\
        $\mathcal{D}$ & Dataset of expert demonstrated trajectories  \\
        $\mathcal{L}(w)$ & Log-likelihood of demonstrated trajectories  \\
        $\Delta L$ & Change in log-likelihood upon adding a candidate state to $C$  \\
        $d_{DKL}$ & Divergence threshold parameter for constraint classification  \\
        \bottomrule
    \end{tabularx}
\end{table}
\section{Constraint Muti-Objective Maximum Entropy IRL} \label{appendix-Proof}
Let $\xi$ denote a trajectory and $f(\xi) \in \mathbb{R}^m$ its associated feature vector encoding $m$ objectives. 
For each expert cluster $k$, preferences over objectives are represented by a weight vector $w_k \in \mathbb{R}^m$.
The trajectory-level reward is defined as
\begin{equation}
R_{w_k}(\xi) = w_k^\top f(\xi).
\end{equation}

We assume the presence of a set of hard constraints $C$ such that any trajectory violating these constraints is considered infeasible.
This is encoded using an indicator function
\begin{equation}
\mathbb{I}^C(\xi) =
\begin{cases}
1, & \text{if } \xi \text{ satisfies all constraints in } C, \\
0, & \text{otherwise}.
\end{cases}
\end{equation}

The goal of Maximum Entropy Inverse Reinforcement Learning is to find a probability distribution $P(\xi \mid C, w_k)$ over feasible trajectories that maximizes entropy while matching the empirical feature expectations of expert demonstrations.
Formally, we solve
\begin{align}
\max_{P(\xi)} \quad
& - \sum_{\xi} P(\xi)\log P(\xi) \\
\text{s.t.} \quad
& \sum_{\xi} P(\xi) = 1, \\
& \sum_{\xi} P(\xi) f(\xi) = \hat{f}, \\
& P(\xi) = 0 \quad \text{if } \mathbb{I}^C(\xi)=0.
\end{align}

Introducing Lagrange multipliers $\lambda$ and $w_k$ for the normalization and feature-matching constraints, the Lagrangian becomes
\begin{equation}
\mathcal{L}
= -\sum_{\xi} P(\xi)\log P(\xi)
+ \lambda\left(\sum_{\xi} P(\xi) - 1\right)
+ w_k^\top\left(\sum_{\xi} P(\xi)f(\xi) - \hat{f}\right).
\end{equation}

Taking the derivative of $\mathcal{L}$ with respect to $P(\xi)$ and setting it to zero yields
\begin{equation}
-\log P(\xi) - 1 + \lambda + w_k^\top f(\xi) = 0.
\end{equation}

Solving for $P(\xi)$ gives
\begin{equation}
P(\xi) = \exp\!\big(w_k^\top f(\xi)\big)\exp(\lambda - 1).
\end{equation}

The normalization constant, or partition function, is defined over feasible trajectories as
\begin{equation}
Z(C, w_k) = \sum_{\xi : \mathbb{I}^C(\xi)=1} \exp\!\big(R_{w_k}(\xi)\big).
\end{equation}

Substituting $\exp(\lambda - 1) = 1 / Z(C, w_k)$ yields the final form:
\begin{equation}
P(\xi \mid C, w_k)
= \frac{1}{Z(C, w_k)} \exp\!\big(R_{w_k}(\xi)\big),
\end{equation}
for all $\xi$ satisfying the constraints, and $P(\xi \mid C, w_k)=0$ otherwise. The idea is taken from Maximum Entropy Inverse Reinforcement Learning by Zeibert et al. 
%%%%%%%%%%%
%%%%%%%%%%%%%%%%%%%%%%%%%%%%%%%
\section{Sensitivity analysis of MOCI}
\label{sec:tuning_ddkl}
The magnitude of the change in log-likelihood, $\Delta L$, upon adding a candidate state to the constraint set $C$ is intrinsically dependent on three environmental factors:
%%%%%%%%%%%%%%%%
\begin{itemize}
    \item \textbf{Dataset Size ($|D|$):} The total log-likelihood $L$ is an unnormalized sum over all trajectories in the expert dataset. Consequently, the magnitude of $\Delta L$ scales linearly with the number of expert demonstrations. 
    \item \textbf{State Space Complexity:} In environments with limited pathways (e.g., a heavily constrained Gridworld), removing a single valid state drastically reduces the partition function $Z$. Conversely, in highly redundant state spaces, the reduction in $Z$ is minimal unless the state represents a critical navigational chokepoint.
    \item \textbf{State Centrality:} Unvisited states located on optimal paths yield a massive $\Delta L$ when constrained, whereas unvisited states in peripheral or suboptimal regions yield negligible changes to the partition function.
\end{itemize}
%%%%%%%%%%%%%%%%
\subsection{Ablation: latent mixture versus a single preference model}
\label{sec:ablation_single_preference}

Heterogeneous demonstrations are modeled as a mixture over $K$ latent linear preferences with shared hard constraints.
A natural baseline is to \emph{remove} mixture capacity by setting $K{=}1$, i.e., to explain the same pooled trajectories with a single MaxEnt component while retaining the remainder of the MOCI procedure (EM updates, constraint search, and the same hyperparameters otherwise).
We instantiate MOCI with $K{\in}\{1,2\}$ on the \emph{same} unlabeled dataset $\mathcal{D}$ for each random demo seed: trajectories are drawn i.i.d.\ from two expert policies with distinct terrain preferences, then pooled and shuffled.
This isolates whether mixture structure materially affects fit and constraint recovery when the generative process truly uses two preference types.

Table~\ref{tab:exp1_k_ablation} summarizes means $\pm$ standard deviations over $50$ independent demo seeds on the heterogeneous GridWorld protocol described above ($d_{\mathrm{DKL}}{=}0.05$, $10$ EM iterations per run).
We report average marginal log-likelihood per trajectory under the fitted model, constraint mean squared error (CMSE), precision and recall for forbidden states, their harmonic mean (F1), and false positive rate (FPR).

\begin{table}[t]
  \centering
  \small
  \caption{Pooled heterogeneous demonstrations: single-preference MOCI ($K{=}1$) versus mixture MOCI ($K{=}2$). Means $\pm$ std.\ over $50$ seeds.}
  \label{tab:exp1_k_ablation}
  \begin{tabular}{lcc}
    \hline
    Metric & $K{=}1$ & $K{=}2$ \\
    \hline
    Avg.\ marginal $\log$-likelihood (per traj.) &
      $-203.732 \pm 0.146$ & $-203.699 \pm 0.156$ \\
    CMSE (constraints) &
      $0.0267 \pm 0.0376$ & $0.0233 \pm 0.0324$ \\
    Precision &
      $0.857 \pm 0.176$ & $0.869 \pm 0.157$ \\
    Recall &
      $0.935 \pm 0.141$ & $0.955 \pm 0.109$ \\
    F1 (forbidden states) &
      $0.891 \pm 0.155$ & $0.907 \pm 0.129$ \\
    FPR &
      $0.0219 \pm 0.0270$ & $0.0206 \pm 0.0257$ \\
    \hline
  \end{tabular}
\end{table}

The ablation supports the hypothesis that mixture structure improves \emph{explanation} of pooled heterogeneous data: $K{=}2$ achieves a systematic, statistically significant lift in marginal likelihood relative to $K{=}1$.
Effects on overlap-based constraint scores are directionally favorable for $K{=}2$ . 
Accordingly, we treat $K{=}1$ primarily as a likelihood-controlled sanity check on mixture capacity rather than as a substitute for the full model in heterogeneous regimes.
%%%%%%%%%%%%%%%%%%%%%%%%%%%%%%%%%
%\input{Sections/Related_Work}
%%%%%%%%%%%%%%%%%%%%%%%%%%%%%%%%%%%%%%%%
\section{MOCI Algorithm}
\label{Sec:em_mlci_algo}
The MOCI algorithm-\ref{alg:em_mlci} employs an Expectation-Maximization (EM) style approach to infer shared constraints and multi-objective rewards jointly. During the E-step, it calculates the responsibility of each reward cluster for the observed expert trajectories. The M-step then updates the cluster priors, optimizes the reward weights for each objective via Maximum Entropy Inverse Reinforcement Learning, and greedily adds candidate constraints from unvisited states as long as they significantly improve the data likelihood without falling below the divergence threshold $d_{DKL}$.
%%%%%%%%%%%%%%%%%%%%%%%%%%
\begin{algorithm}
\small
\caption{Multi-Objective Constraint Inference (MOCI)}
\label{alg:em_mlci}
\SetAlgoLined
\DontPrintSemicolon % Hides semicolons at the end of lines for a cleaner, pythonic look
% Elegant Input/Output formatting
\KwIn{MDP $\mathcal{M}$, Demonstrations $\mathcal{D} = \{\xi_1, \dots, \xi_N\}$, Number of clusters $K$, threshold $d_{DKL}$, Priors $\{\pi_k\}_{k=1}^K$, Learning rate $\alpha$}
\KwOut{Inferred constraints $C$, Reward weights $\{w_k\}_{k=1}^K$}
\vspace{1.5mm}
%\hrule
\vspace{1.5mm}
\textbf{Initialization:}\\
$C \leftarrow \emptyset$\;
$\mathcal{C}_{cand} \leftarrow \{s \in \mathcal{S} \mid s \notin \xi \text{ for all } \xi \in \mathcal{D}\}$ \tcp*{Unvisited states}
Initialize $w_k$ randomly, $\pi_k \leftarrow 1/K$ for $k \in \{1, \dots, K\}$\;
\vspace{2mm}

\Repeat{convergence of $\mathcal{L}(C, \{w_k\}, \{\pi_k\})$}{
    \vspace{1.5mm}
    \tcc{\textbf{E-Step: Evaluate Responsibilities}}
    \For{$k = 1$ \KwTo $K$}{
        Compute partition function $Z(C, w_k)$ via backward pass\;
    }
    \For{$i = 1$ \KwTo $N$}{
        \For{$k = 1$ \KwTo $K$}{
            $\gamma_{i,k} \leftarrow \frac{\pi_k P(\xi_i \mid C, w_k)}{\sum_{j=1}^K \pi_j P(\xi_i \mid C, w_j)}$\;
        }
    }
    \vspace{2mm}
    
    \tcc{\textbf{M-Step A: Update Priors}}
    \For{$k = 1$ \KwTo $K$}{
        $\pi_k \leftarrow \frac{1}{N} \sum_{i=1}^N \gamma_{i,k}$\;
    }
    \vspace{2mm}
    
    \tcc{\textbf{M-Step B: Update Reward Weights (MaxEnt IRL)}}
    \For{$k = 1$ \KwTo $K$}{
        $\nabla_{w_k} \mathcal{L} \leftarrow \sum_{i=1}^N \gamma_{i,k} \left( \phi(\xi_i) - \mathbb{E}_{P(\xi \mid C, w_k)}[\phi(\xi)] \right)$\;
        $w_k \leftarrow w_k + \alpha \nabla_{w_k} \mathcal{L}$ %\tcp*{Gradient Ascent}
    }
    \vspace{2mm}
    
    \tcc{\textbf{M-Step C: Update Shared Constraints}}
    \While{$\mathcal{C}_{cand} \neq \emptyset$}{
        $c^* \leftarrow \arg\max_{c \in \mathcal{C}_{cand}} \mathcal{L}(C \cup \{c\}, \{w_k\}, \{\pi_k\})$\;
        $\Delta \mathcal{L} \leftarrow \mathcal{L}(C \cup \{c^*\}, \cdot) - \mathcal{L}(C, \cdot)$\;
        \eIf{$\Delta \mathcal{L} \le d_{DKL}$}{
            \textbf{break} \tcp*{Stop adding constraints to prevent overfitting}
        }{
            $C \leftarrow C \cup \{\hat{c}\}$\;
            $\mathcal{C}_{cand} \leftarrow \mathcal{C}_{cand} \setminus \{\hat{c}\}$\;
        }
    }
}
\end{algorithm}